# Improving TSP Solutions Using GA with a New Hybrid Mutation Based on Knowledge and Randomness

Esra'a Alkafaween, Ahmad B. A. Hassanat
IT Department, Mutah University, Mutah, Karak, Jordan

## Abstract

Genetic algorithm (GA) is an efficient tool for solving optimization problems by evolving solutions, as it mimics the Darwinian theory of natural evolution. The mutation operator is one of the key success factors in GA, as it is considered the exploration operator of GA.

Various mutation operators exist to solve hard combinatorial problems such as the TSP. In this paper, we propose a hybrid mutation operator called "IRGIBNNM", this mutation is a combination of two existing mutations; a knowledge-based mutation, and a random-based mutation. We also improve the existing "select best mutation" strategy using the proposed mutation.

We conducted several experiments on twelve benchmark Symmetric traveling salesman problem (STSP) instances. The results of our experiments show the efficiency of the proposed mutation, particularly when we use it with some other mutations.

**Keyword:** Knowledge-based mutation, Inversion mutation, Slide mutation, RGIBNNM, SBM.

## 1. Introduction

### 1.1 Travelling salesman problem (TSP)

TSP is considered as one of the combinatorial optimization problems [1], that is easy to describe but difficult to solve, and it is classified among the problems that are not solved in polynomial time; i.e. it belongs to the NP-hard problem [2].

A solution of TSP aims at finding the shortest path (tour) through a set of nodes (starting from a node N and finishing at the same node) so that each node is visited only once [3].

The classic problem of traveling salesman is an active and attractive field of research because of its simple formulation [2], and it was proved to be NP-complete problem, since no one found any effective way to solve an NP problem of a large size, in addition, many problems in the world can be modeled by TSP [4].

The TSP is classified into:

1. Symmetric traveling salesman problem (STSP):
The cost (distance) between any two cities in both directions is the same (undirected graph), i.e. the distance from *city1* to *city2* is the same as the distance from *city2* to *city1*. There are (N-1)! /2 possible solutions for N cities.

2. Asymmetric travelling salesman problem (ATSP): The cost between any two cities in both directions is not the same. There are (N-1)! possible solutions for N cities [5].

TSPs are used in various applications, including : job sequencing, Computer wiring, Crystallography, Wallpaper cutting, Dartboard design, Hole punching, Overhauling gas turbine, etc. [6].

Over the years various techniques have been suggested to solve the TSP, such as Genetic Algorithm (GA) [7] [8], Hill Climbing [9], Nearest Neighbor and Minimum Spanning Tree algorithms [10], Simulated Annealing [11], Ant Colony [9], Tabu Search [12], Particle Swarm [13], Elastic Nets [14], Neural Networks [15], etc. Genetic algorithms are one of the algorithms that extensively applied to solve the TSP [16].

## 1.2 Genetic Algorithm (GA)

GA is an optimization algorithm [17] that is classified as global search heuristic; it is one of the categories that form the family of the evolutionary algorithms, which mimics the principles of natural evolution [18]. GA is a population-based search algorithm, as in each generation, a new population is generated by repeating three basic operations on the population, namely, selection, crossover, and mutation [19]. GA has been used extensively in many fields, such as computer networks [20], speech recognition [21], image processing [22], software engineering [23], etc.

A simple GA algorithm is described as follows [16]:

**Step1:** Create a random population of potential solutions [24] consisting of *n* individuals (initial populations)**.**
**Step2:** Evaluate the fitness value f(x) of each individual, x, in the population.
**Step3**: Repeat the following three steps to create a new population until completion of the new population.
**Step4:** Select two individuals of the current generation for mating.
**Step5**: Apply crossover with a certain ratio to create offspring.
**Step6**: Apply mutation with a certain ratio.
**Step7**: The previous operations are repeated until the completion criterion is met.

The performance of the GA is affected by several key factors, such as the population size, the selection's strategy, the mutation operator used, the crossover operator used and the coding scheme [25], [26]. In this paper, we focus on the mutation operator.

Mutation operator plays an important role in the GA, where it helps to stimulate the GA to explore new areas in the search space [19]. It is an effective mechanism for preserving the diversity of individuals [25], where mutation provides variation in the population through random changes of individuals [26]. And therefore, overcoming the so-called premature convergence [27], also preventing the loss of genetic material [28].

In this paper, we propose a hybrid mutation operator called inversion RGIBNNM (IRGIBNNM) to provide an efficient solution to TSP, we use simple GA with mutations only; there is no other variable/parameter that controls the workflow of such a simple GA, as we want to examine the strength of the proposed mutation apart from the effect of other parameters; we compare the performance of this mutation with the performances of three existing mutations, and we used it with two other mutations to form a multi-mutations GA. The comparisons are made on symmetric TSP instances.

The organization of this paper is as follows. In Section 2 we present some of the related work. In section 3, we present the proposed mutation, the existing two mutations and the mutation strategy. In Section 4 we describe the experiments conducted, and discuss the. The conclusion is written in Section 5.

## 2. Related work

Over the years, researchers have suggested several types of mutations to be used in various types of encoding, including: Flip Mutation, Creep Mutation and insert mutation [29], Gaussian Mutation, Exchange Mutation [30], Displacement Mutation [31], Uniform Mutation [1], Inversion Mutation [32] and some other types.

Louis and Tang proposed a new mutation called Greedy-swap mutation, where two cities are chosen randomly from the same chromosome, and switching between them if the length of the new tour obtained is "better" (shorter) than the previous ones [33].

Potvin [2] and Larrañaga et al. [8] presented a review of representing the TSP, explaining the advantages and disadvantages of different mutation operators. Soni and Kumar studied many types of mutations that solve the problem of travelling salesmen, including Interchanging Mutation, Reversing Mutation and Scramble Mutation [1]. Otman and Jaafar used Reverse Sequence Mutation (RSM) and several types of crossover to solve the TSP [28]. Korejo et al, introduced a directed mutation (DM), this method used the statistical information provided by the current population to explore the promising areas in the search space [19].

Having such a large number of mutations, the problem becomes which mutation to use? As the problem lies in choosing the appropriate mutation. To answer this question, several researchers have developed new types of GAs that use more than one mutation at the same time [34], [35], [36] and [37].

Katayama et al. presented a promising GA for TSP, called a hybrid mutation genetic algorithm (HMGA), which employed a local search algorithm called stochastic hill climbing (SHC), in order to avoid falling into the local optima [38]. Hong et al. proposed a new GA, called dynamic genetic algorithm (DGA) in order to choose the appropriate mutation and crossover operators and their ratios automatically, this algorithm use more than one mutation at the same time, such as: Uniform crossover, (0, 1) change, Inversion, Bit-change and Swapping [39].

Hassanat et al. proposed 10 types of knowledge-based mutations; the most promising one is called "Random Gene Inserted beside nearest neighbor mutation" (RGIBNNM). In addition, they proposed two selection strategies for the mutation operators called: "select the best mutation" (SBM) and "select any mutation" (SAM). They applied all mutations and strategies on several TSP instances [34].

Regardless the extensive research in this domain, there is no one mutation ideally suited for all TSP instances. Since no one method found in the literature that guarantees an optimal solution for any TSP instance. Therefore, there is still room for improvement in this domain.

## 3. The proposed method

In this section, we explain some of the existing mutation operators that are proposed for the permutation coded GA; these include slide mutation [40], inversion mutation [32] and RGIBNNM [34]. Moreover, we explore the strategy of choosing the best mutation; the SBM [34]. We also present the proposed hybrid mutation, which is nothing but a combination of the inversion mutation and the RGIBNNM, we call it IRGIBNNM.

### 3.1 Slide mutation

This mutation chooses two genes randomly, and then conveys the first to follow the second, and then shift the rest of the city, as depicted by Example (1).

**Example 1:** Consider the following TSP tour (C):

$$C = (5\ 3\ \mathbf{10}\ 2\ 1\ 8\ 9\ \mathbf{7}\ 4\ 6).$$

If the third gene (10) and the eighth gene (7) are randomly selected, then the sub tour is:

(**2 1 8 9 7**). The mutated tour will be: (Offspring) = (**5 3 10 1 8 9 7 2 4 6**).

### 3.2 Inversion mutation

This mutation chooses two random genes, and then reverses the subset between them, as depicted by Example (2).

**Example 2:** Consider the following tour (C):

C= (**5 3 10 2 1 8 9 7 4 6**).

If the third and eighth positions are randomly selected, then the sub tour is (**2 1 8 9 7**), and then reversed to be (**7 9 8 1 2**).

The mutated tour will be: (Offspring) = (**5 3 10 7 9 8 1 2 4 6**).

## 3.3 RGIBNNM mutation

This mutation is a knowledge-based operator designed especially for the TSP problem. However, it can be customized to fit some other problems. This operator uses the idea of the nearest neighbor cities, where this mutation selects a random gene (city), and finds its nearest city, then swap the random city with one of the neighbors of the nearest city.

## 3.4 The proposed IRGIBNNM

We propose a hybrid mutation called: IRGIBNNM.

In this mutation we combine two mutation operators, the inversion mutation and RGIBNNM.

The IRGIBNNM initially applies the inversion mutation on an individual, and then the RGIBNNM is applied to the resulting individual. Thus, the new offspring benefit from both mutations' characteristics, attempting to enhance the performance of both mutations, by increasing diversity in the search space, and therefore to provide better results. The IRGIBNNM is depicted by Example (4).

**Example 4:** Consider the following tour (C):

C= (**5 3 10 9 8 1 2 7 4**) with cost =**19**, as depicted in Figure (2). To apply IRGIBNNM:

1. Select two random genes, e.g. the third and eighth genes.
2. A= Inversion Mutation(C). The resulting offspring
   A = (**5 3 10 2 1 8 9 7 4**) with cost = **18.2** (see Figure 2).
3. Apply RGIBNNM(A) as follows:
- Select a random gene from A, e.g. the random gene is the eighth gene, i.e. the random city is (**7**).
- Find the nearest city to the random city (7), which is city (3) in this case.
- Get a random city around city (3) in the range (± 5); e.g. city (**9**).
- Apply the Exchange mutation on chromosome A by swapping the cities 7 and 9, as shown in (Figure (3)). The final output offspring becomes:
  Offspring = (**5 3 10 2 1 8 7 9 4**) with cost = (**17.1**).

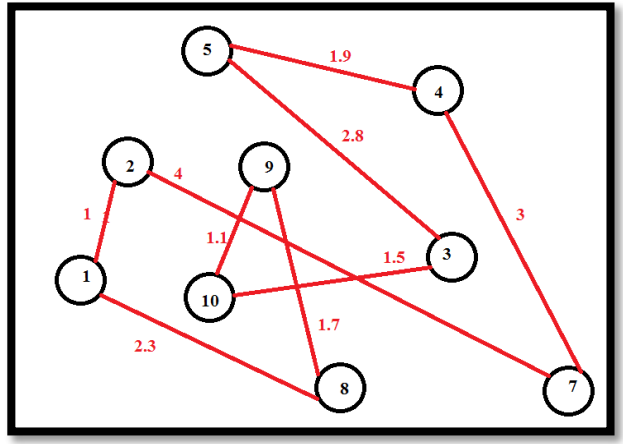

Figure 1. Example of particular tour (C) with cost=19.

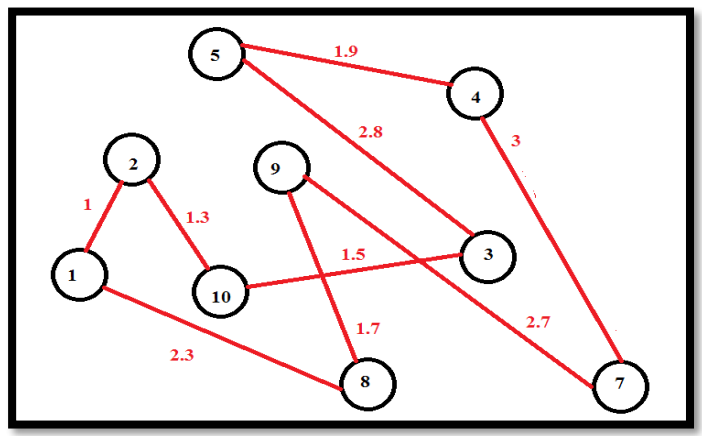

Figure 2. Example of applying Inversion mutation on C to get offspring A with cost=18.2.

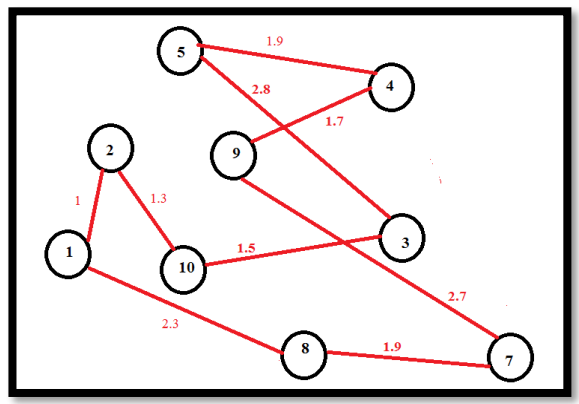

Figure 3. Example of applying IRGIBNNM on A to get offspring with cost=17.1..

## 3.5 Select the best mutation (SBM)

This strategy applies various mutation operators simultaneously to the same individual, and from each mutation produces one offspring; the "best" offspring that does not already exist in the population is added to the population [34]. For TSP the "best" solution, is the one with the minimum rate.

In this paper, we used three mutations only (Slide mutation, Inversion mutation, and the proposed IRGIBNNM), instead using several other mutations as proposed by [34].

A larger example is shown in Figure (4), which depicts the implementation of four mutations, in addition to the SBM strategy for 80 random cities.

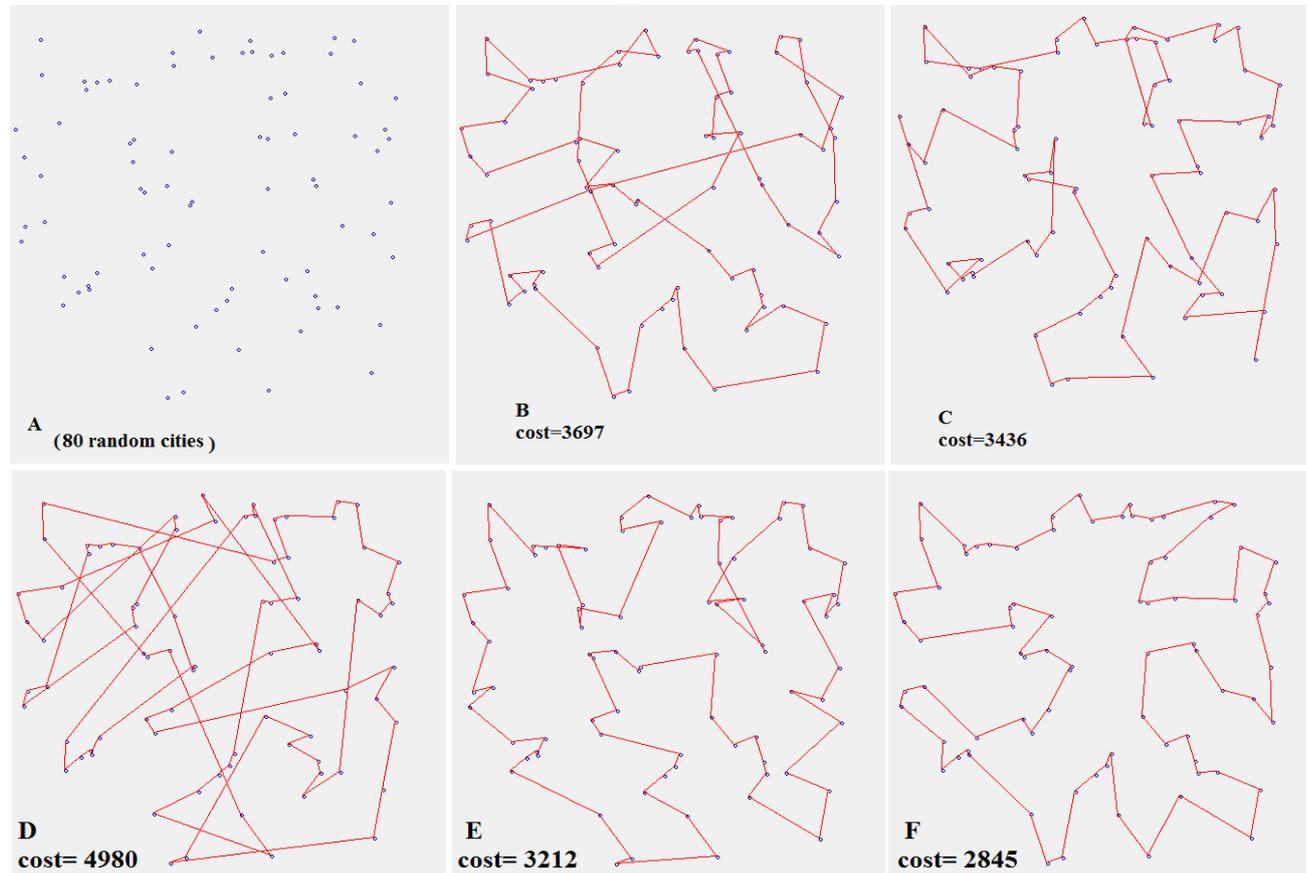

**Figure 4. Implementation of the four mutations, (A) 80 Random cities, (B)Slide mutation, (C) inversion mutation, (D) RGIBNNM, (E) IRGIBNNM, (F) SBM strategy.**

A real data example is shown in Figure (5), which shows the implementation of the four mutations and SBM on a particular route of the TSP (eil51) taken from TSBLIB [41] . A closer look at Figures (4 and 5) shows significant improvements on the initial tour, particularly when using IRGIBNNM or SBM strategy.

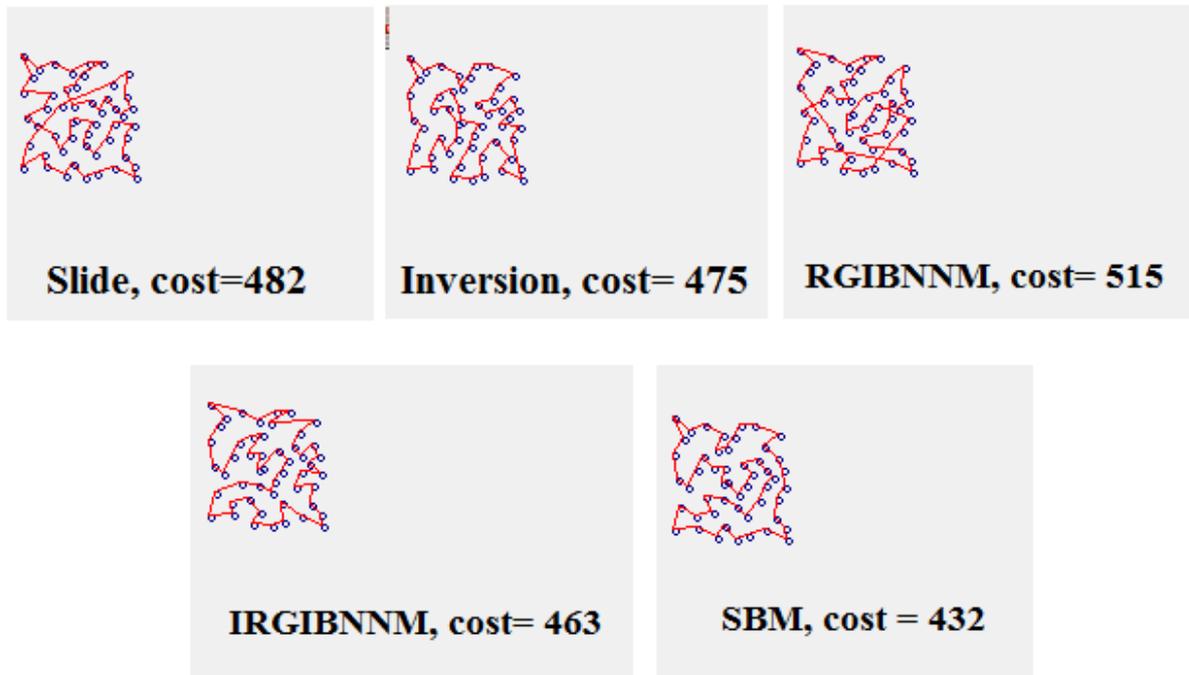

Figure 5.The effect of applying the mutations on Eil51.

# 4.Experimental setup and Result

To evaluate the performance of the proposed mutation (IRGIBNNM) and the new SBM strategy, we conducted several experiments on twelve TSP instances, each having the best known solution (optimal). Those instances were taken from the TSPLIB [41], and they include: eil51, a280, bier127, berlin52, KroA100, KroA200, ch150, rat195, st70, pr125, pr226 and lin318. Same experiments, under the same circumstances were conducted to examine the convergence to a minimum value of each operator separately, including the other mutations (slide, inversion)

We have implemented the new SBM strategy the same as proposed by [34], but using three mutations only (slide, inversion, and the proposed IRGIBNNM), instead of several other mutations, considering the best offspring to be added to the population. To prevent duplication of chromosomes, if the new offspring is found in the population, we consider the lower quality offspring, and if all of the three offspring found in the population the operation (on that chromosome) is canceled.

In all experiments, our GA used the reinsertion method, which is an expansion sampling [42], where this method means, only the excellent half (from the new individuals and old generation) is selected as a population for the next generation.

in other words, when creating a new generation, the old generation competes with the new individuals.

We repeated each experiment 10 times, the GA parameters used are as follows: the Population size = 100, the probability of crossover = 0% and all previous mutations occur 100%. The initial population is random based population seeding and selection strategy in all algorithms is random. The termination criterion is based on a fixed number of generations reached. For all of our experiments the maximum number of generations = 2000.

The operators are coded in VC++, and the computer specifications: 1.66 GHz processor PC with 2 GB of RAM.

The results of the mutations evaluated on 12 instances from the TSP are summarized in Table (1 and 2).

**Table 1. Results of TSP instances obtained by Inversion and Slide mutations after 2000 generations.**

| Mutation type | | Inversion mutation | | | Slide mutation | | |
|---|---|---|---|---|---|---|---|
| Instances | Optimal | Best Fitness | Worst fitness | Average fitness | Best Fitness | Worst fitness | Average fitness |
| **eil51** | 426 | **440** | 453 | 446.1 | 469 | 583 | 503.9 |
| **a280** | 2579 | 9811 | 10119 | 9974.2 | 9532 | 10522 | 9917.4 |
| **bier127** | 118282 | 167565 | 183857 | 172867.4 | 177720 | 193326 | 185276.6 |
| **kroA100** | 21282 | 30310 | 33413 | 31925.3 | 31800 | 36279 | 34120.6 |
| **berlin52** | 7542 | **7769** | 8515 | 8038.1 | 8498 | 10154 | 9334.6 |
| **kroA200** | 29368 | 80906 | 84555 | 81958 | 74586 | 90348 | 83529.8 |
| **pr125** | 73682 | 151643 | 168468 | 161445.4 | 170304 | 218119 | 192498 |
| **lin318** | 42029 | 185852 | 192611 | 188931.6 | 176935 | 185899 | 181978.5 |
| **pr226** | 80369 | 331572 | 353613 | 342094.3 | 345027 | 377088 | 360239.9 |
| **ch150** | 6528 | 13006 | 13670 | 13425.1 | 13129 | 15221 | 13778.1 |
| **st70** | 675 | 758 | 815 | 783 | 787 | 1004 | 882.4 |
| **rat195** | 2323 | 5548 | 5955 | 5836.5 | 5420 | 6169 | 5774.8 |

**Table 2. Results of TSP instances obtained by IRGIBNNM and RGIBNNM mutations after 2000 generations.**

| Mutation type | | IRGIBNNM | | | RGIBNNM | | |
|---|---|---|---|---|---|---|---|
| Instances | Optimal | Best Fitness | Worst fitness | Average fitness | Best Fitness | Worst fitness | Average fitness |
| **eil51** | 426 | 448 | 463 | 455.3 | 518 | 603 | 575.5 |
| **a280** | 2579 | **7313** | 7846 | 7507.9 | 6543 | 8307 | 7526.5 |
| **bier127** | 118282 | **156903** | 169657 | 164072.9 | 205820 | 254541 | 234760.2 |
| **kroA100** | 21282 | **25941** | 29218 | 27418.7 | 43474 | 53903 | 48077.1 |
| **berlin52** | 7542 | 8098 | 8705 | 8354.2 | 9639 | 11105 | 10296.1 |

| kroA200 | 29368 | **59802** | 63911 | 62136.9 | 88409 | 109892 | 97125.7 |
| pr125 | 73682 | **111055** | 127783 | 121013.5 | 213526 | 270814 | 235064.1 |
| lin318 | 42029 | **132899** | 145109 | 136569.5 | 159856 | 178241 | 173127.6 |
| pr226 | 80369 | **191049** | 234720 | 216699 | 288421 | 380900 | 322855.1 |
| ch150 | 6528 | **10517** | 11396 | 11111.9 | 15071 | 18435 | 16774.2 |
| st70 | 675 | **733** | 772 | 753.4 | 1058 | 1296 | 1222.1 |
| rat195 | 2323 | **4321** | 4758 | 4554.2 | 6203 | 7492 | 7081.5 |

As can be seen from Tables (1 and 2), the best performance was recorded by the IRGIBNNM for 10 instances, followed by the Inversion mutation, which also shows a better performance than both of the Slide mutation and the RGIBNNM. The significant performance of the IRGIBNNM is justified by the exploiting of two mutations applied after each other on the same individual. The first provides random solutions and the second provides solution based on some knowledge of the nearest neighbor. Randomness provided by the inversion mutation, and knowledge provided by the RGIBNNM allow for more diversity of good solutions, which leads to better results.

Figure (6) shows the convergence to the minimum value recorded by each mutation. Again IRGIBNNM shows faster convergence to the minimum value than the other two mutations on KroA100. This faster convergence is due to the same randomness and knowledge afforded by the IRGIBNNM.

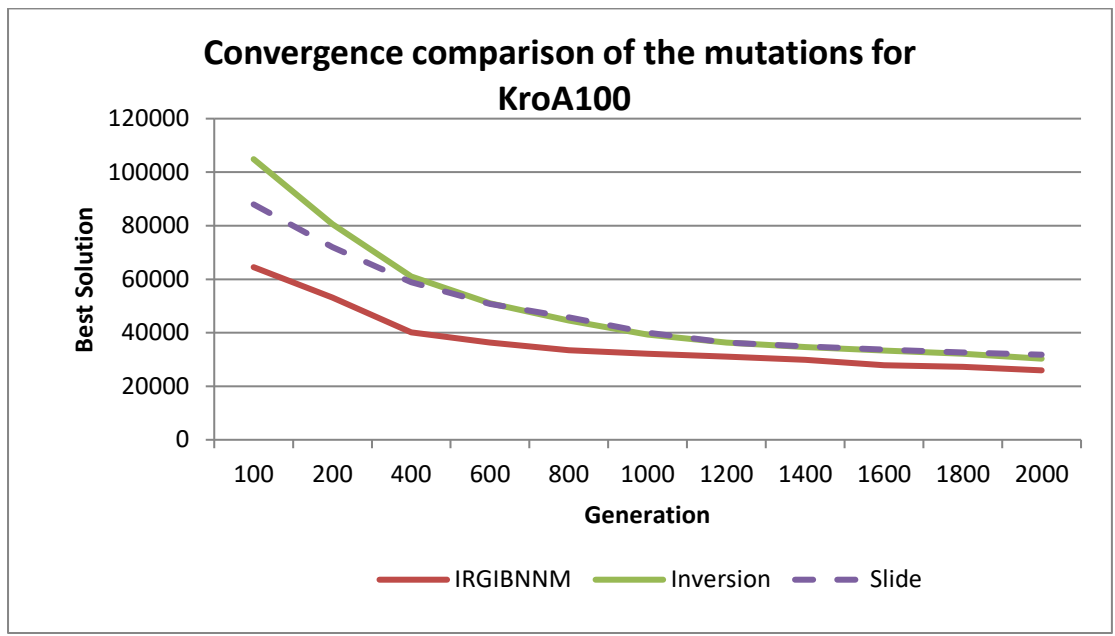

**Figure 6. Mutation's convergence to the minimum solution, kroA100.**

Using the same GA parameters, the second set of experiments is conducted to measure the performance of the new SBM, and to show the effective use of more than one mutation at the same time by the GAs. The results are shown in Table (3) and Figures (7, 8 and 9).

Table 3. Results of TSP instances obtained by SBM after 2000 generations

| Instances | Optimal | Best Fitness | Worst fitness | Average fitness |
|---|---|---|---|---|
| **eil51** | 426 | **428** | 439 | 432.7 |
| **a280** | 2579 | **2898** | 3089 | 2974.9 |
| **bier127** | 118282 | **121644** | 128562 | 124492.5 |
| **kroA100** | 21282 | **21344** | 22788 | 21957.1 |
| **berlin52** | 7542 | **7544** | 8423 | 7890.7 |
| **kroA200** | 29368 | **30344** | 32103 | 31369 |
| **pr152** | 73682 | **74777** | 86240 | 77022.9 |
| **lin318** | 42029 | **47006** | 50033 | 48234.6 |
| **pr226** | 80369 | **82579** | 87006 | 84409.1 |
| **ch150** | 6528 | **6737** | 7044 | 6876 |
| **st70** | 675 | **677** | 723 | 694.8 |
| **rat195** | 2323 | **2404** | 2561 | 2481.9 |

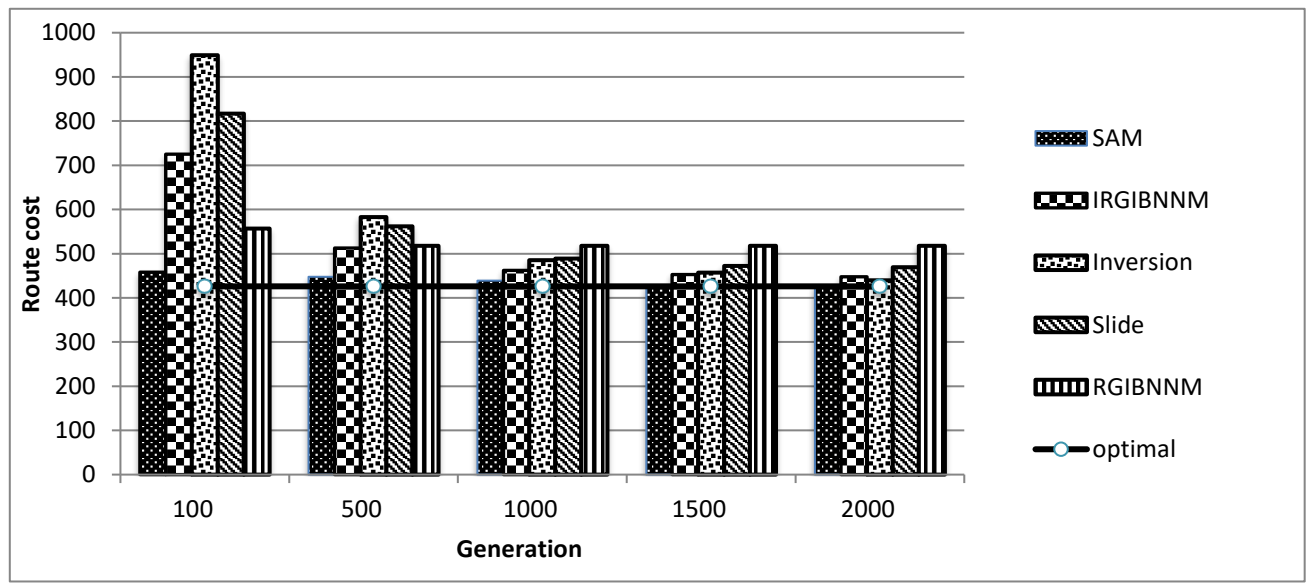

Figure 7. Convergence Comparison for eil51.

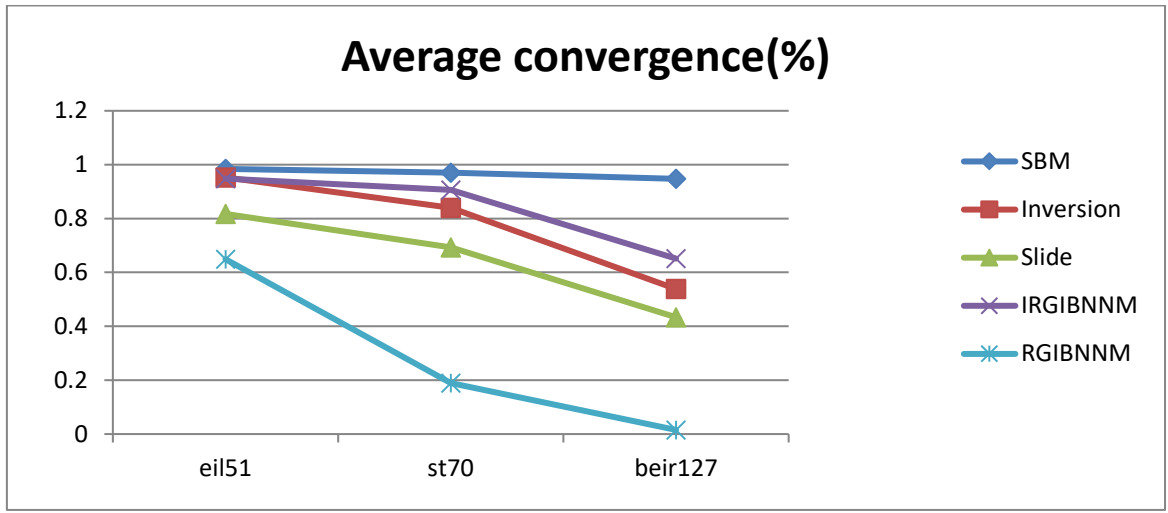

Figure 8. Average Convergence of 4 mutations and SBM strategy for three instances from TSPLIB (eil51, st70, beir127).

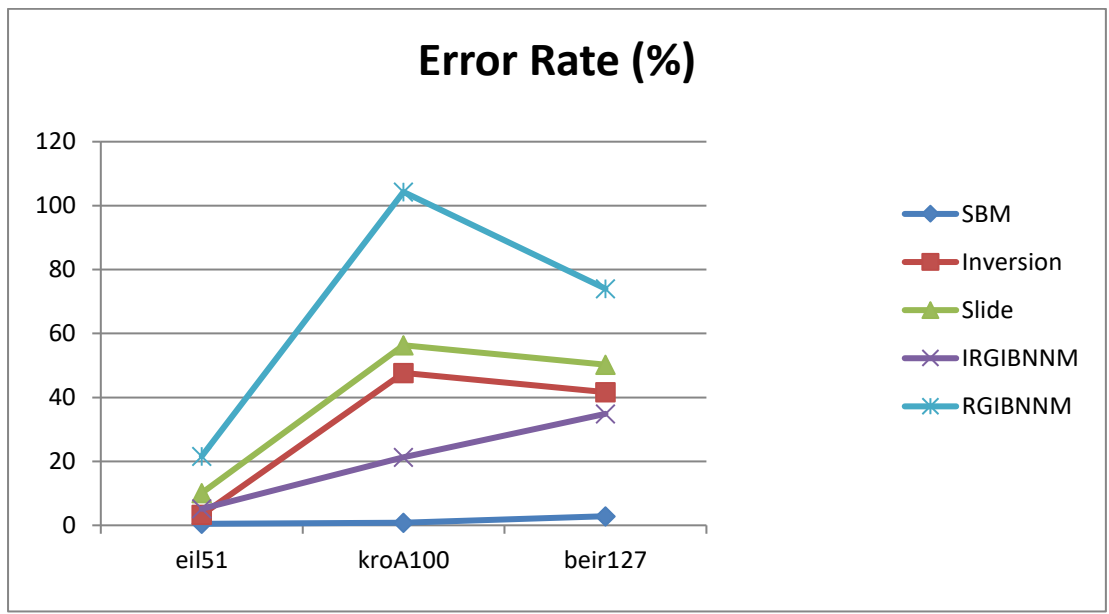

Figure 9. Error rate of 4 mutations and SBM strategy for three instances from TSPLIB (eil51, st70, beir127).

As can be seen from the results, it is important to select the appropriate mutation, in particular for the SBM strategy, and in general for the GA, because the choice of those methods affects the results of the GA significantly. As seen in Table (3), the best performance was recorded by the SBM algorithm, followed by the proposed IRGIBNNM.

As seen in Figure 7, the SBM performs better than the other mutations, it is interesting to note that the solutions provided by the SBM are very close to the optimal solutions for most of the TSP instances examined. Results from Figures (8 and 9) shows the efficient use the three mutations together by the SBM, where the SBM achieves the highest convergence and less error with significant difference.

We justify the significant performance of the SBM as follows, intuitively, we have 2 options for the quality of a solution provided by any mutation, comparing to the average quality in the current population, a) a lower quality solution, and b) a higher quality solution; assuming that a solution with the same quality is considered as a higher quality solution. The new SBM uses 3 mutations, which are applied on the same chromosome, the probability to have them all fail, (i.e. to get lower quality outcomes (offspring) from all mutations used) is 1 out of 8 (low, low, low), while the probability to get a higher quality by any of them is 7 out of 8 possibilities, this high success rate justifies the significant performance of the SBM. Same justification applies to the good performance of the proposed IRGIBNNM, but with a lower success rate of 3 out of 4, since the IRGIBNNM uses only 2 mutations .

The success of the new SBM is not attributed only to the use of multi mutations as described above, but also to the quality of the solutions provided by the mutations used by the SBM, and this pays attention to the proposed IRGIBNNM, which used by the SBM among the other two mutations. This conclusion is supported by comparing the results of the new SBM with the old SBM proposed by [34], see Table (4).

Table 4. Results of new SBM compared to those of old SBM.

| Instances | Optimal | New SBM 2000 Generations | Old SBM [34] 1600 generations |
| --- | --- | --- | --- |
| eil51 | 426 | 428 | 30706 |
| a280 | 2579 | 2898 | 100975 |
| bier127 | 118282 | 121644 | 53541 |
| kroA100 | 21282 | 21344 | 48506 |
| berlin52 | 7542 | 7544 | 29452 |
| kroA200 | 29368 | 30344 | - |
| pr152 | 73682 | 74777 | - |
| lin318 | 42029 | 47006 | - |
| pr226 | 80369 | 82579 | - |
| ch150 | 6528 | 6737 | - |
| st70 | 675 | 677 | - |
| rat195 | 2323 | 2404 | - |

The parameters of the GA proposed by [34] are almost the same except for the number of generations, in this paper, we used 2000 iterations, while in [34] they used 1600, but this would not bias the comparison made in Table (4), because of the very large gap in the quality of the solutions compared, since an extra 400 iterations cannot bridge such a large gap in the performance of the old SBM, neither gives a huge favor to the new SBM.

Comparing the proposed methods with the plethora of mutations found in the literature is not appreciated, because of the different parameters used by different GAs, such as the number of generations, the mutation rate, crossover rate, population size, selection method, initial population seeding, etc., since each of these parameters affects the results of the GA significantly.

Time complexity for most of mutations found in the literature designed for the TSP ranges from $O(1)$ (such as the simple-random-swapping algorithms) to $O(N)$ (for more complex mutations such as the Slide, Inversion and RGIBNNM mutations, where N is the number of cities in a TSP instance.

The time complexity of the proposed IRGIBNNM mutation is $O(2N)$, since it uses two mutations of order N. Accordingly, the Time complexity of the new SBM is $O(4N)$ comparing to the old SBM, which has $O(10N)$ as it uses ten $O(N)$ mutations. Asymptotically both of the proposed methods are of $O(N)$, but in practice they definitely consume more time than most of the mutations found in the literature. Surprisingly, both algorithms might be used to speed up the GA; this is due to their fast convergence to a minimum solution. See Figure (6 and 7), using just the first 100 iterations the GA converged to high quality solutions.

# 5. Conclusion

In this paper, we propose a hybrid mutation based on knowledge of the TSP and random swapping) called "IRGIBNNM" to enhance the performance of the GA for solving the TSP. We have compared the performance of the IRGIBNNM with three existing mutations, in addition to the SBM, which in this work used three mutations including the proposed one.

The experimental results of 12 TSP instances show the efficiency of the proposed mutation, and the strength of the new SBM, both of the proposed methods benefit from randomness and knowledge provided by the nearest neighbor approach. Also, both methods benefit from the increased probability of getting new high quality solutions due to the use of more than one mutation.

The high quality solutions for the TSP obtained by a GA, which used only the mutation operator, without using other advanced options that used by state-of-the-art GA such as advanced crossovers, initial seeding, advanced selection methods, adaptive change of population size and mutation/crossover rates, etc. The future work will focus on employing the proposed method with other advanced operators to further enhance the performance of the GA when applied for solving the TSP.